\documentclass{article}

\PassOptionsToPackage{sort&compress,numbers}{natbib}

\usepackage[final]{neurips_2021}

\usepackage[utf8]{inputenc} 
\usepackage[T1]{fontenc}    
\usepackage{hyperref}       
\usepackage{url}            
\usepackage{booktabs}       
\usepackage{amsfonts}       
\usepackage{nicefrac}       
\usepackage{microtype}      
\usepackage{xcolor}         
\usepackage{grffile,xspace,multirow}
\usepackage{amsmath}
\usepackage[ruled,vlined]{algorithm2e}
\usepackage{wrapfig}

\usepackage{enumitem}
\setlist{noitemsep}
\setlist[1]{noitemsep}
\setlist[1]{nosep}

\newlist{compactitem}{itemize}{3}
\setlist[compactitem]{topsep=0pt,partopsep=0pt,itemsep=0pt,parsep=0pt,leftmargin=*}
\setlist[compactitem,1]{label=\textbullet}
\setlist[compactitem,2]{label=---}
\setlist[compactitem,3]{label=*}

\newlist{compactdesc}{description}{3}
\setlist[compactdesc]{topsep=0pt,partopsep=0pt,itemsep=0pt,parsep=0pt}

\newlist{compactenum}{enumerate}{3}
\setlist[compactenum]{topsep=0pt,partopsep=0pt,itemsep=0pt,parsep=0pt}
\setlist[compactenum,1]{label=\arabic*}
\setlist[compactenum,2]{label=\alph*}
\setlist[compactenum,3]{label=\roman*}

\newcommand{\argmin}{\operatornamewithlimits{argmin}}

\def\Input{\textbf{Input}}
\def\Return{\textbf{Return}}
\def\yb{{\mathbf{y}}}
\def\xb{{\mathbf{x}}}
\def\vb{{\mathbf{v}}}
\newcommand{\thetab}{{\boldsymbol{\theta}}}
\newcommand{\wb}{{\mathbf{w}}}
\newcommand{\Wb}{{\mathbf{W}}}
\newcommand{\Pb}{{\mathbf{P}}}
\newcommand{\Cb}{{\mathbf{C}}}
\newcommand{\Zb}{{\mathbf{Z}}}
\newcommand{\Yb}{{\mathbf{Y}}}
\newcommand{\Xb}{{\mathbf{X}}}
\newcommand{\Mb}{{\mathbf{M}}}
\newcommand{\Rb}{{\mathbf{R}}}
\newcommand{\R}{\mathbb{R}}

\newcommand{\best}[1]{\bf #1}

\newcommand{\liblinear}{LIBLINEAR\xspace}
\newcommand{\xmc}{XMC\xspace}
\newcommand{\xmclong}{eXtreme Multi-label Text Classification \xspace}

\newcommand{\xtransformer}{{X-Transformer}\xspace}
\newcommand{\xrlinear}{{XR-Linear}\xspace}
\newcommand{\xrtransformer}{{XR-Transformer}\xspace}
\newcommand{\pretrained}{{Pre-Trained}\xspace}
\newcommand{\wordvec}{{word2vec}\xspace}

\newcommand{\tfidf}{TF-IDF\xspace}

\newcommand{\attentionxml}{AttentionXML\xspace}
\newcommand{\annexml}{AnnexML\xspace}
\newcommand{\bonsai}{Bonsai\xspace}
\newcommand{\discmec}{DiSMEC\xspace}

\newcommand{\parabel}{Parabel\xspace}
\newcommand{\pfastrexml}{PfastreXML\xspace}

\newcommand{\ppdsparse}{PPD-Sparse\xspace}
\newcommand{\proxml}{ProXML\xspace}

\newcommand{\slice}{SLICE\xspace}
\newcommand{\xmlcnn}{XML-CNN\xspace}
\newcommand{\xtext}{eXtremeText\xspace}
\newcommand{\napkinxc}{NAPKINXC\xspace}
\newcommand{\xreg}{XReg\xspace}
\newcommand{\glas}{GLaS\xspace}
\newcommand{\hnsw}{HNSW\xspace}
\newcommand{\scann}{ScaNN\xspace}
\newcommand{\mach}{MACH\xspace}
\newcommand{\deepxml}{DeepXML\xspace}
\newcommand{\decaf}{DECAF\xspace}
\newcommand{\galaxc}{GalaXC\xspace}
\newcommand{\eclare}{ECLARE\xspace}
\newcommand{\aplcxlnet}{APLC-XLNet\xspace}
\newcommand{\lightxml}{LightXML\xspace}

\newcommand{\eurlex}{{\sf Eurlex-4K}\xspace}
\newcommand{\wikis}{{\sf Wiki10-31K}\xspace}
\newcommand{\amzcat}{{\sf AmazonCat-13K}\xspace}
\newcommand{\wikil}{{\sf Wiki-500K}\xspace}
\newcommand{\amzsmall}{{\sf Amazon-670K}\xspace}
\newcommand{\amzlarge}{{\sf Amazon-3M}\xspace}

\title{Fast Multi-Resolution Transformer Fine-tuning for Extreme Multi-label
Text Classification}

\author{%
  Jiong Zhang\\
  Amazon\\
  \footnotesize{\texttt{jiongz@amazon.com}}
  \And
  Wei-cheng Chang\\
  Amazon\\
  \footnotesize{\texttt{chanweic@amazon.com}}
  \And
  Hsiang-fu Yu\\
  Amazon\\
  \footnotesize{\texttt{rofu.yu@gmail.com}}
  \And
  Inderjit S. Dhillon\\
  UT Austin \& Amazon\\
  \footnotesize{\texttt{inderjit@cs.utexas.edu}}
}

\begin{document}

\maketitle

\begin{abstract}
  Extreme multi-label text classification~(\xmc) seeks to find
  relevant labels from an extreme large label collection for a given text
  input. Many real-world applications can be formulated as \xmc problems, such
  as recommendation systems, document tagging and semantic search. Recently,
  transformer based \xmc methods, such as \xtransformer and \lightxml,
  have shown significant improvement over other \xmc methods.
  Despite leveraging pre-trained transformer models for text representation,
  the fine-tuning procedure of transformer models on large label space
  still has lengthy computational time even with powerful GPUs.
  In this paper, we propose a novel \textit{recursive} approach, \xrtransformer to accelerate
  the procedure through recursively fine-tuning transformer models on a series
  of multi-resolution objectives related to the original \xmc objective function.
  Empirical results show that \xrtransformer takes significantly less
  training time compared to other transformer-based \xmc models while yielding
  better state-of-the-art results. In particular, on the public \amzlarge dataset with 3
  million labels, \xrtransformer is not only 20x faster than \xtransformer but
  also improves the Precision@1 from $51\%$ to $54\%$.
  Our code is publicly available at \url{https://github.com/amzn/pecos}.
\end{abstract}

\vspace{-.5em}
\section{Introduction}
\vspace{-.5em}

Many real-world applications such as open-domain question
answering~\cite{chang2020pretraining,lee2019latent}, e-commerce dynamic search
advertising~\cite{prabhu2018parabel,prabhu2014fastxml}, and semantic matching~\cite{chang2021semantic}, can be formulated as an
extreme multi-label text classification~(\xmc) problem: given a text input,
predict relevant labels from an enormous label collection of size $L$.
In these applications, $L$ ranges from tens of thousands to millions, which
makes it very challenging to design \xmc models that are
both accurate and efficient to train. Recent works such as
\parabel~\citep{prabhu2018parabel}, \bonsai~\citep{khandagale2020bonsai},
\xrlinear~\citep{yu2020pecos} and \attentionxml~\cite{you2019attentionxml},
exploit the correlations among the labels to generate label partitions or
hierarchical label trees~(HLTs) which can be used to shortlist candidate
labels to be considered during training and inference. While these methods are scalable in
terms of the size of the label collection, most of them rely only on statistical
representations (such as bag-of-words) or pooling from pre-generated token
embeddings (such as word2vec) to vectorize text inputs.

In light of the recent success of deep pretrained transformers models such as
BERT~\cite{devlin2018bert}, XLNet~\cite{yang2019xlnet} and
RoBerta~\cite{liu2019roberta} in various NLP applications,
\xtransformer~\cite{chang2020xmctransformer} and
\lightxml~\cite{jiang2021lightxml} propose to fine-tune pre-trained
transformer models on \xmc tasks to obtain new state-of-the-art results over the
aforementioned approaches. Although transformers are able to better capture
semantic meaning of textual inputs than statistical representations, text
truncation is often needed in practice to reduce GPU memory footprint and
maintain model efficiency. For example, \xtransformer truncates input texts to
contain the first 128 tokens before feeding it into transformer models.
Efficiency of transformer fine-tuning poses another challenge for \xmc
applications. Directly fine-tuning transformer models on the
original \xmc task with a very large label collection is infeasible as both
the training time and the memory consumption are linear in $L$. In order to
alleviate this, both \xtransformer and \lightxml adopt a similar approach to
group $L$ labels into $K$ clusters of roughly equal size denoted by $B$ and
fine-tune transformers on the task to identify relevant label clusters (instead of
labels themselves). If $B\approx\sqrt{L}$ and $K \approx \sqrt{L}$, then both the
training time and the memory requirement of the fine-tuning can be reduced to
$O(\sqrt{L})$ from $O(L)$.  However, as pointed out
in~\cite{you2019attentionxml}, the model performance would deteriorate due to
the information loss from label aggregation.
Thus, both \xtransformer and \lightxml still choose a small constant $B$ (
$\le 100$) as the size of the label clusters. As a result, transformers are
still fine-tuned on a task with around $L/100$ clusters, which leads to
a much longer training time compared with non-transformer based models. For
example, it takes \xtransformer 23 and 25 days respectively to train on
\amzlarge and \wikil even with 8 Nvidia V100 GPUs.

To address these issues, we propose \xrtransformer, an \xmc architecture that
leverages pre-trained transformer models and has much smaller training cost
compared to other transformer-based \xmc methods. Motivated by the
multi-resolution learning in image
generation~\cite{lai2017deep,karras2017progressive,karras2019style} and
curriculum learning~\cite{bengio2009curriculum}, we formulate the \xmc problem
as a series of sub-problems with multi-resolution label signals and recursively
fine-tune the pre-trained transformer on the coarse-to-fine objectives.
In this paper, our contributions are as follows:
\begin{itemize}
  \item We propose \xrtransformer, a transformer based framework for extreme multi-label text
    classification where the pre-trained transformer is recursively fine-tuned
    on a series of easy-to-hard training objectives defined by a hierarchical
    label tree.  This allows the transformers to be quickly fine-tuned for a
    \xmc problem with a very large number label collection progressively.
  \item To get better text representation and mitigate the information loss in
    text truncation for transformers, we take into account statistical text
    features in addition to the transformer text embeddings in our model.
    Also, a cost sensitive learning scheme by label aggregation is proposed to
    introduce richer information on the coarsified labels.
  \item We conduct experiments on 6 public XMC benchmarking datasets and our
    model takes significantly lower training time compared to other
    transformer-based XMC models to yield better state-of-the-art results. For
    example, we improve the state-of-the-art Prec@1 result on Amazon-3M established by
    \xtransformer from 51.20\% to 54.04\% while reducing
    the required training time from 23 days to 29 hours using the same hardware.
\end{itemize}

\vspace{-0.5em}
\section{Related Works}
\vspace{-0.5em}
\paragraph{Sparse Linear Models with Partitioning Techniques.}
Conventional \xmc methods consider fixed input representations
such as sparse \tfidf features and study different partitioning techniques
or surrogate loss functions on the large output spaces to reduce complexity.
For example, sparse linear one-versus-all (OVA) methods such as
\discmec~\citep{babbar2017dismec},
\ppdsparse~\citep{yen2016pd,yen2017ppdsparse},
\proxml~\citep{babbar2019data}
explore parallelism to solve OVA losses
and reduce the model size by weight truncations.

The inference time complexity of OVA models is linear in the output space,
which can be greatly improved by partitioning methods
or approximate nearest neighbor (ANN) indexing on the label spaces.
Initial works on tree-based methods~\citep{choromanska2015logarithmic, 
daume2017logarithmic} reduce the OVA problem to one-versus-some (OVS) with logarithmic depth trees.  
Down that path, recent works on sparse linear models including
\parabel~\citep{prabhu2018parabel},
\xtext~\citep{wydmuch2018no},
\bonsai~\citep{khandagale2020bonsai},
\xreg~\citep{prabhu2020extreme},
\napkinxc~\citep{jasinska2020probabilistic,jasinska2021online} and
\xrlinear~\citep{yu2020pecos}
partition labels with $B$-array hierarchical label trees (HLT),
leading to inference time complexity that is logarithmic in the output space.
On the other hand, low-dimensional embedding-based models
often leverage ANN methods to speed up the inference procedure.
For example,
\annexml~\citep{tagami2017annexml} and
\slice~\citep{jain2019slice}
consider graph-based methods such as \hnsw~\citep{malkov2020hnsw}
while \glas~\citep{guo2019breaking}
considers product quantization variants such as \scann~\citep{guo2020accelerating}.

\paragraph{Shallow Embedding-based Methods.}
Neural-based \xmc models employ various network architectures to learn semantic
embeddings of the input text.
\xmlcnn~\citep{liu2017deep} applies one-dimensional CNN on the input sequence
and use the BCE loss without sampling, which is not scalable to \xmc problems.
\attentionxml~\citep{you2019attentionxml} employs BiLSTMs and label-aware attention
as scoring functions. For better scalability to large output spaces,
only a small number of positive and hard negative labels
are used in model GPU training.
Shallow embedding-based methods
~\citep{medini2019extreme,dahiya2021deepxml,mittal2021decaf,saini2021galaxc,mittal2021eclare}
use word embedding lookup followed by shallow MLP layers to obtain input embeddings.
For instance, \mach~\citep{medini2019extreme} learns MLP layers on
several smaller \xmc sub-problems induced by hashing tricks on the large label space.
Similarly, \deepxml~\citep{dahiya2021deepxml} and its variant
(i.e., \decaf~\citep{mittal2021decaf}, \galaxc~\citep{saini2021galaxc}, \eclare~\citep{mittal2021eclare})
pre-train MLP encoders on \xmc sub-problems induced by label clusters.
They freeze the pre-trained word embedding and learn another MLP layer
followed by a linear ranker with sampled hard negative labels from \hnsw~\citep{malkov2020hnsw}.
Importantly, shallow embedding-based methods only show competitive performance
on short-text \xmc problems where the number of input tokens is small
~\citep{medini2019extreme,dahiya2021deepxml}.

\paragraph{Deep Transformer Models.}
Recently, pre-trained Transformer models~\citep{devlin2018bert,liu2019roberta,yang2019xlnet}
have been applied to \xmc problems with promising results
~\citep{chang2020xmctransformer,ye2020pretrained,jiang2021lightxml}.
\xtransformer~\citep{chang2020xmctransformer} considers a two-stage approach
where the first stage transformer-based encoders are learned on \xmc sub-problems
induced by balanced label clusters, and the second stage
sparse \tfidf is combined with the learned neural embeddings as the input
to linear OVA models.
\aplcxlnet~\citep{ye2020pretrained} fine-tunes XLNet encoder on adaptive imbalanced label
clusters based on label frequency similar to Adaptive Softmax~\citep{joulin2017efficient}.
\lightxml~\citep{jiang2021lightxml} fine-tunes Transformer encoders with the OVA loss
function end-to-end via dynamic negative sampling from the matching network trained on
label cluster signals.
Nonetheless, Transformer-based \xmc models have larger model size and
require longer training time, which hinders its practical usage on different downstream
\xmc problems.

\section{Background Material}
We assume we are given a training set $\{\xb_i, \yb_i\}_{i=1}^N$ where $\xb_i\in\mathcal{D}$ is
the $i$th input document and $\yb_i\in\{0,1\}^L$ is the one hot label vector
with $y_{i,\ell}=1$ indicating that label $\ell$ is relevant to instance $i$.
The goal of \xmclong (\xmc) is to learn a function $f: \mathcal{D}\times [L]\mapsto \R$, such that
$f(\xb, \ell)$ denotes the relevance between the input $\xb$ and the label $\ell$. In
practice, labels with the largest $k$ values are retrieved as the predicted
relevant labels for a given input $\xb$. The most straightforward model is
one-versus-all~(OVA) model:
\begin{align}
	f(\xb, \ell) = \wb_{\ell}^\top \Phi(\xb); \ \ell \in[L], \label{eq:ova}
\end{align}
where $\Wb=[\wb_1,\ldots,\wb_L]\in\R^{d\times L}$ are the weight vectors
and $\Phi: \mathcal{D}\mapsto \R^d$ is the text vectorizer that maps $\xb$
to $d$-dimensional feature vector. $\Phi(\cdot)$ could be a deterministic text
vectorizer, such as the bag-of-words~(BOW) model or Term Frequency-Inverse
Document Frequency~(TFIDF) model, or a vectorizer with learnable parameters.
With the recent development in deep learning, using pre-trained transformer as
the text vectorizer has shown promising results in many \xmc
applications~\cite{ye2020pretrained,chang2020xmctransformer,jiang2021lightxml}.
When $L$ is large, however, training and inference of OVA model without sampling
would be prohibitive due to the $O(L)$ time complexity.

To handle the extremely large output space, recent approaches partition
the label space to shortlist the labels considered during training and inference.
In particular,
~\citep{medini2019extreme,chang2020xmctransformer,yu2020pecos,ye2020pretrained,dahiya2021deepxml,jiang2021lightxml}
follow a three stage framework: partitioning, shortlisting, and ranking.
First, label features are constructed to group labels into $K$ clusters $\Cb\in\{0,1\}^{L\times K}$ where
$C_{\ell,k}=1$ denotes that label $\ell$ is included in the $k$-th cluster.
Then a shortlisting model is learned to match input $\xb$ to relevant clusters
in an OVA setting:
\begin{align}
  g(\xb, k) = \hat{\wb}_k^\top \Phi_g(\xb);\>k\in[K].
  \label{eq:matcher}
\end{align}
Finally, a classification model with output size $L$ is trained on the shortlisted labels:
\begin{align}
	f(\xb, \ell) = \wb_{\ell}^\top \Phi(\xb); \ \ell \in S_g(\xb),
  \label{eq:ranker}
\end{align}
where $S_g(\xb)\subset[L]$ is the label set shortlisted by $g(\xb,\cdot)$.
In the extreme case where only one label cluster is determined to be
relevant to a input $\xb$, the training and inference cost on $\xb$ would be
$O(K+\frac{L}{K})$, which in the best case scenario is $O(\sqrt{L})$ when $K=\sqrt{L}$.

For transformer based methods, the dominant time is the evaluation of
$\Phi(\xb)$.
But $K$ being too big or too small could still be problematic.
Empirical results show that the model performance deteriorates when
clusters are too big~\cite{you2019attentionxml}. This is because that the signals coming from $B$ labels
within the same cluster will be aggregated and not distinguishable, where $B$
is the cluster size.
Therefore, $B$ cannot be too big to ensure a reasonable
label resolution for fine-tuning.
Also, as pointed out in~\cite{chang2020xmctransformer}, fine-tuning transformer
models on large output spaces can be prohibitive.
As a result,
the label clusters need to be constructed in a way to balance the model
performance and fine-tuning efficiency.
In practice, both the transformer-based XMC models, such as \xtransformer and
\lightxml, adopt a small fixed constant as the cluster size $B$ ($\le 100$), which means that
training the shortlisting model $g(\xb, k)$ is still very time consuming as
the number of clusters $K\approx L/B$.

\section{Proposed Method: \xrtransformer}
As noted above, the shortlisting problem~\eqref{eq:matcher}
is itself an \xmc problem with slightly smaller output size $\frac{L}{B}$ where
$B$ is the cluster size.
In \xrtransformer, we apply the same three stage framework recursively on the
shortlisting problem until a reasonably small output size is reached
$\frac{L}{B^D}$.
We can therefore follow the curriculum learning scheme
and fine-tune the pre-trained transformers progressively on the sub-\xmc
problems with increasing output space $\{\frac{L}{B^D},\frac{L}{B^{D-1}},\ldots\}$.
At each fine-tuning task, the candidate label set is shortlisted by the final
model at the previous task. The recursive shortlisting ensures that for any input,
the number of candidate labels to include in training and inference is $O(B)$
and therefore the total number of considered labels is $O(B\log_{B}(L))$.
Also, we leverage the multi-step fine-tuning and use the embedding generated
at the previous task to bootstrap the non pre-trained part for the current task.
We now describe the model design in detail.


\begin{figure}[t]
\centering
\includegraphics[width=1.0\textwidth]{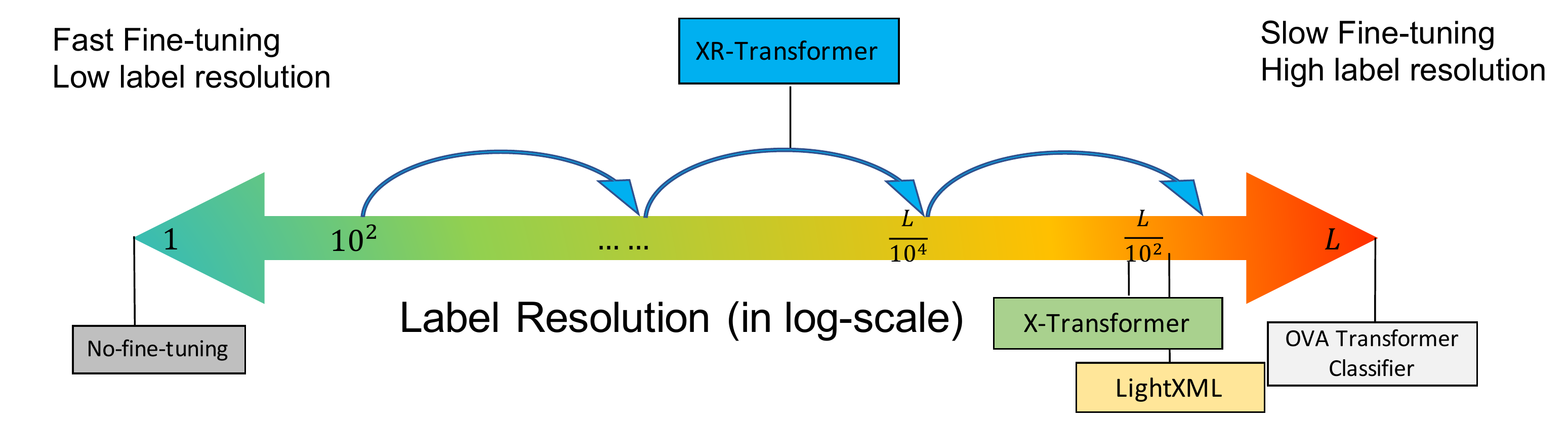}
\caption{Illustration of fine-tuning on \xmc tasks of different label resolutions.
  For an \xmc task with a low label resolution, fine-tuning can be fast but
  model performance might deteriorate due to large deviation from the
  original \xmc task. In practice, \xtransformer and
  \lightxml adopt a \xmc task with a relatively higher label resolution to ensure
  reasonable model performance at the cost of longer training time.
  The proposed \xrtransformer leverages
  multi-resolution learning and model bootstrapping that achieves both fast
  fine-tuning and good model performance.
  }
  \label{fig:multi-resolution}
\end{figure}

\paragraph{Hierarchical Label Tree~(HLT).}
Recursively generating label clusters $D$ times is equivalent to building a
HLT~\cite{jain2016extreme} of depth $D$. We first
construct label features $\Zb\in\R^{L\times \hat{d}}$.
This could be done by applying text vectorizers on label text or from
Positive Instance Feature Aggregation~(PIFA):
\begin{align}
  \Zb_{\ell} &= \frac{\vb_\ell}{\|\vb_\ell\|};\text{where } \vb_\ell=\sum_{i:y_{i,\ell}=1}
  \Phi(\xb_i), \forall\>\ell\in[L],\label{eq:pifa}
\end{align}
where $\Phi:\mathcal{D}\mapsto \R^{d}$ it the text vectorizer.
Then we follow similar procedures
as~\cite{you2019attentionxml} and~\cite{jiang2021lightxml} and use
balanced k-means~($k=B$) to recursively partition label sets and generate the HLT in a
top-down fashion. The HLT is represented with a series of indexing matrices
$\{\Cb^{(t)}\}_{t=1}^D$, such that $\Cb^{(t)}\in\{0,1\}^{K_t\times K_{t-1}}$
where $K_0=1$ and $K_D=L$. Equivalently, once $\Cb^{(D)}$ is constructed, the
HLT can be built from bottom up through joining $B$ adjacent clusters together.

\paragraph{Multi-resolution Output Space.}
Multi-resolution learning has been explored in different contexts
such as computer vision~\cite{pedersoli2015coarse,lai2017deep}.
For instance, using an output scheme with coarse-to-fine resolutions results in
better image quality for generative adversarial networks~\cite{karras2017progressive,karras2019style}.
As an another example in meta learning,
~\citep{liu2019self} learns multiclass models via auxiliary meta classes
by collapsing existing classes.
Nevertheless, multi-resolution learning has not been well-explored
in the \xmc literature.
In \xrtransformer, we leverage
the label hierarchy defined by the HLT and train the transformer model on
multi-resolution objectives.

The \xmc task can be viewed as generating an 1-D image
$\yb\in\{0,1\}^{L}$ with binary values based on input text $\xb$.
Just like a coarsified image could be obtained by a max or mean pooling of
nearby pixels, the coarse label vector can be obtained by max-pooling of
labels which are nearby in the label feature space.
Once the HLT is constructed using label features, the true labels
at layer $\Yb^{(t)}\in\{0,1\}^{N\times K_t}$ can be determined by the
true labels of the child clusters at $t+1$ through a max-pooling like
operation:
\begin{align}
  \Yb^{(t)}=\text{binarize}(\Yb^{(t+1)}\Cb^{(t+1)}),\label{eq:label-tree}
\end{align}
and $Y^{(D)}_{i,\ell}=y_{i,\ell}$ is the original label matrix. This forms a
series of learning signals with coarse-to-fine resolution and can be used to
generate learning tasks with easy-to-hard objectives.

Direct use of the binarized $\Yb^{(t)} \in \{0,1\}^{N \times K_t}$ in Eq~\eqref{eq:label-tree}
results in information loss when merging several positive labels into one cluster.
Ideally, a cluster containing several positive children is more relevant than
a cluster with only one positive child.
To add this lower level information to higher level learning objectives,
we introduce the relevance matrix $\Rb^{(t)} \in \R_{+}^{N \times K_t}$ for
layer $t$ of the \xmc sub-problem where
$R_{i,\ell}$ defines the non-negative important weight for $i$th instance to $\ell$th cluster.
Different from cost-sensitive learning~\citep{elkan2001foundations} for MLC
(CSMLC) setting~\citep{li2014condensed,huang2017cost,lin2019advances}
where there is only one cost matrix explicitly derived by evaluation
metrics such as F1 score,
in \xrtransformer, we consider the usage of cost-sensitive learning where
the relevance matrices are recursively induced by the HLT structure.
Specifically, given an HLT, we recursively construct relevance matrices for $t=1,\ldots,D$:
\begin{align}
\Rb^{(t)} = \Rb^{(t+1)}\Cb^{(t+1)},
\end{align}
and $\Rb^{(D)}=\Yb^{(D)}$.
Motivated by~\citep{menon2019multilabel},
we adopts the row-wise $\ell_1$ normalized relevance matrix:
  \[
    \hat{R}^{(t)}_{i,j}=\left\{
                \begin{array}{ll}
                  \frac{R^{(t)}_{i,j}}{\|\Rb^{(t)}_{i}\|_1} &\text{ if }Y^{(t)}_{i,j}=1,\\
                  \alpha &\text{otherwise,}
                \end{array}
              \right.
  \]
where $\alpha$ is the hyper parameter to balance positive and negative
weights.

\paragraph{Label Shortlisting.}
During training, \xrtransformer only focuses on discriminating the
labels or clusters that have high chance of being positive.
A necessary condition for a label at layer $t$ to
be positive is that its parent label at level $t-1$ is positive.
Therefore, an intuitive approach would be to only train on the output space
shortlisted by positive clusters of the parent layer. However, in practice we
found this approach sometimes leads to sub-optimal result during inference with
beam search. As an effort to balance explore and exploit, we further include
the top-$k$ relevant clusters determined by the model learned on the parent
layer to mimic the beam search during inference. Thus at layer $t$, the labels
considered during training are shortlisted by the parent layer $t-1$:
\begin{align}
  \Pb^{(t-1)} &= \text{Top}(\Wb^{(t-1)\top}\Phi(\Xb,\Theta^{(t-1)}), k),\label{eq:topk_pred}\\
  \Mb^{(t)} &= \text{binarize}(\Pb^{(t-1)}\Cb^{(t)\top}) + \text{binarize}(\Yb^{(t-1)}\Cb^{(t)\top}),\label{eq:shortlist}
\end{align}
where the $\text{Top}(\cdot, k)$ operator zeros out elements in a matrix except the
top-$k$ largest values in each row. For each instance $\xb_i$, only non-zero
indices of $\Mb_i$ will be included into the training objective.
We can therefore define a series of learning objectives for level
$t\in\{1,2,\ldots,D\}$ as:
\begin{align}
  \min_{\Wb^{(t)}, \Theta} \sum_{i=1}^N
  \sum_{\ell:\Mb^{(t)}_{i,\ell}\neq 0} \hat{R}^{(t)}_{i,\ell}\mathcal{L}(Y^{(t)}_{i,\ell}, \Wb^{(t)\top}_\ell
  \Phi(\xb_i, \Theta)) + \lambda \|\Wb^{(t)}\|^2,\label{eq:obj}
\end{align}
where $\mathcal{L}$ is a point-wise loss such as hinge loss, squared hinge
loss or BCE loss, $\Wb^{(t)}$, $\Theta$ are the model weights to be
learned.

\paragraph{Text Representation.}
Most previous works on XMC construct text feature representation in one of
two ways: statistical feature representations and semantic feature
representations. Although the latter, in particular transformer models, have
shown promising results on various NLP benchmarks, the self-attention mechanism makes transformers
unscalable w.r.t. sequence length. To ensure efficiency, input texts are
usually truncated~\cite{chang2020xmctransformer,jiang2021lightxml} which
result in loss of information. On the other hand, the statistical features,
such as TFIDF, are fast to construct with the whole input taken into
consideration.
In \xrtransformer, we use a combination of these two feature
representations and each component is a complement of lost information for the
other one:
\begin{align}
  \Phi_{cat}(\xb,\Theta) := \Big[ \frac{\Phi_{tfidf}(\xb)}{\|\Phi_{tfidf}(\xb)\|}, \frac{\Phi_{dnn}(\xb, \Theta)}{\|\Phi_{dnn}(\xb, \Theta)\|} \Big], \label{eq:vectorizer}
\end{align}
where $\Phi_{dnn}(\cdot,\Theta)$ is the transformer parametrized by $\Theta$. Once
the text representation is constructed, predictions can be made by simply
applying a linear projection on top of the text representation through~\eqref{eq:ova}.

\begin{algorithm}[t]\label{alg:iterative_learn}
  \SetKwInOut{Input}{Input}
  \SetKwInOut{Return}{Return}
  \Input{$\Xb$, $\Yb$, $\Cb$, $\thetab$, $\Pb$}
  $m,n\leftarrow \Cb.shape$

  \eIf{$n=1$}{
    $\Mb\leftarrow ones(N, m)$

    \eIf{$\thetab$ is not fixed}{
      $\thetab^*\leftarrow$ Optimize~\eqref{eq:obj} with $\Phi=\Phi_{dnn}$,
      initialize $\Theta=\thetab$
    }{
     $\thetab^*\leftarrow \theta$
    }
     $\Wb^* \leftarrow \argmin_{\Wb} \sum_{i=1}^N
     \sum_{\ell=1}^m \hat{R}^{(t)}_{i,\ell}\mathcal{L}(Y_{i,\ell}, \Wb^\top_\ell \Phi_{cat}(\xb_i, \Theta)) + \lambda \|\Wb\|^2
     $
   }{
     $\Yb_{prev}\leftarrow \text{binarize}(\Yb\Cb)$

     $\Mb\leftarrow \text{binarize}(\Pb\Cb^\top) + \text{binarize}(\Yb_{prev}\Cb^\top)$

    \eIf{$\thetab$ is not fixed}{
      $\thetab^*\leftarrow$ Optimize~\eqref{eq:obj} with $\Phi=\Phi_{dnn}$ initialize $\Theta=\thetab$
    }{
     $\thetab^*\leftarrow \thetab$
    }
    $\Wb^* \leftarrow \argmin_{\Wb} \sum_{i=1}^N
    \sum_{\ell:\Mb_{i,\ell}\neq 0} \hat{R}^{(t)}_{i,\ell}\mathcal{L}(Y_{i,\ell}, \Wb^\top_\ell
    \Phi_{cat}(\xb_i, \theta^*))  + \lambda \|\Wb\|^2
    $
  }
  \Return{$\Wb^*$, $\thetab^*$}

  \caption{$\text{Iterative\_Learn}(\Xb, \Yb, \Cb, \thetab, \Pb)$}
\end{algorithm}

\begin{algorithm}[t]\label{alg:xrtransformer_train_iter}
  \SetKwInOut{Input}{Input}
  \SetKwInOut{Return}{Return}
  \Input{$\Xb$, $\Yb$, pre-trained transformer $\Phi_{dnn}(\cdot,\thetab_0)$}
  $\hat{\Zb}_\ell = \vb_\ell / \|\vb_\ell\|; \text{where } \vb_{\ell}=\sum_{i:y_{i,\ell}=1}
  \Phi_{tfidf}(\xb_i), \forall\>\ell\in[L]$

  $\{\hat{\Cb}^{(t)}\}_{t=1}^{\hat{D}} \leftarrow \text{k-means-clustering}(\hat{\Zb})$

  Generate label hierarchy $\{\hat{\Yb}^{(t)}\}_{t=1}^{\hat{D}}$ using~\ref{eq:label-tree}

  $\thetab^*=\thetab_0$, $\Pb=None$

  \For{$t$ in $1,2,3,\cdots,\hat{D}$}{
    $\hat{\Wb}, \thetab^* \leftarrow \text{Iterative\_Learn}(\Xb, \hat{\Yb}^{(t)}, \hat{\Cb}^{(t)}, \thetab^*, \Pb)$

  $\Pb \leftarrow \text{Top}(\hat{\Wb}^{\top}\Phi_{cat}(\Xb,\thetab^*), k)$
  }

  $\Zb_\ell = \vb_\ell / \|\vb_\ell\|; \text{where } \vb_\ell=\sum_{i:y_{i,\ell}=1}
  \Phi_{cat}(\xb_i), \forall\>\ell\in[L]$

  $\{\Cb^{(t)}\}_{t=1}^D \leftarrow \text{k-means-clustering}(\Zb)$
  
  Generate label hierarchy $\{\Yb^{(t)}\}_{t=1}^{\hat{D}}$ using~\ref{eq:label-tree}

  Fix $\thetab^*$, $\Pb=None$

  \For{$t$ in $1,2,3,\cdots,D$}{
    $\Wb^{(t)}, \_ \leftarrow \text{Iterative\_Learn}(\Xb, \Yb^{(t)}, \Cb^{(t)}, \thetab^*, \Pb)$

  $\Pb \leftarrow \text{Top}(\Wb^{(t)\top}\Phi_{cat}(\Xb,\thetab^*), k)$
  }

  \Return{$\Phi_{cat}(\cdot, \thetab^*)$, $\{\Cb^{(t)}\}_{t=1}^D$, $\{\Wb^{(t)}\}_{t=1}^D$}
  \caption{\xrtransformer training}
\end{algorithm}

\paragraph{Training with bootstrapping.}
The training of \xrtransformer consists of three steps. At first, a preliminary
HLT is constructed using raw statistical features. Then a pre-trained transformer model is
fine-tuned recursively from low resolution output to high resolution. At each
layer $t$, fine-tuning objective~\eqref{eq:obj} is optimized with
initialization $\Theta=\thetab^{(t-1)*}$ the best transformer weights of layer
$t-1$. $\thetab^{(0)*}$ denotes the pre-trained transformer weights.

Unlike the transformer warmed-up with pre-trained weights, the projection
weights $\Wb^{(t)}$ is trained from scratch without good initialization.
At the beginning of fine-tuning, gradient flow through these
cold-start~(usually randomly initialized) weights will usually worsen the
pre-trained components.
We leverage the recursive learning structure to tackle this issue by model
bootstrapping.
Concretely, $\Wb^{(t)}$ is initialized as:
\begin{align}
  \Wb^{(t)}_{init}:=\argmin_{\Wb^{(t)}} \sum_{i=1}^N
  \sum_{\ell:\Mb^{(t)}_{i,\ell}\neq 0} \hat{R}^{(t)}_{i,\ell}\mathcal{L}(Y^{(t)}_{i,\ell}, \Wb^{(t)\top}_\ell
  \Phi_{dnn}(\xb_i, \thetab^{(t-1)*}))+ \lambda \|\Wb^{(t)}\|^2,\label{eq:obj_bootstrapping}
\end{align}
In practice, \eqref{eq:obj_bootstrapping} is fast to compute since the
semantic text feature for the previous layer $\Phi_{cat}(\Xb, \thetab^{(t-1)*})$
is already computed and thus~\eqref{eq:obj_bootstrapping} can be solved very
quickly on CPUs with a variety of parallel linear solvers, such as
\liblinear~\cite{fan2008liblinear}.

Once the fine-tuning is complete, the refined HLT is constructed with the text
representation that combines statistical text feature and fine-tuned semantic
text embeddings. Then the ranking models are trained on top of the combined
text features for the final prediction. The detailed training procedure is
described in Algorithm~\ref{alg:iterative_learn} and \ref{alg:xrtransformer_train_iter}.

\paragraph{Inference.}
The inference cost of \xrtransformer consists mainly of two parts: cost to
compute transformer embedding and to retrieve relevant labels through beam
search. Therefore, the inference time complexity is $O(T_{dnn} + kd\log(L))$,
where $k$ is the beam size, $d$ is the concatenated feature dimension and
$T_{dnn}$ is the time to compute $\Phi_{dnn}(\xb)$ for a given input. Note
that even the inference is done with beam search through the refined HLT, the
transformer text embedding only need to be computed once per instance.

\paragraph{Connections with other tree based methods.}
Although methods such as  
\attentionxml~\cite{you2019attentionxml} also train on supervisions induced by 
label trees, the final model is a chain of sub-models which each on is 
learned on single-resolution. In particular, given a hierarchical label tree 
with depth $D$, \attentionxml will train $D$ different text encoders on each layer of 
the tree where as \xrtransformer trains the same transformer encoder progressively on 
all layers of the tree. This difference leads to a longer inference time for 
\attentionxml than \xrtransformer since multiple text encoders need to be 
queried during inference, as shown in the comparison in the inference time in 
Appendix~\ref{apx:inference}.

\section{Experimental Results}\label{sec:exp}
We evaluate \xrtransformer on 6 public XMC benchmarking datasets:
\eurlex, \wikis, \amzcat, \wikil, \amzsmall, \amzlarge. Data statistics
are given in Table~\ref{table:data}. For fair comparison, we use the
same raw text input, sparse feature representations and same train-test
split as \attentionxml~\citep{you2019attentionxml} and
other latest works~\cite{chang2020xmctransformer,jiang2021lightxml}.
The evaluation metric is Precision@k (P@k), which is widely-used in \xmc literature
~\citep{tagami2017annexml,babbar2017dismec,prabhu2018parabel,you2019attentionxml,chang2020xmctransformer,jiang2021lightxml}.
The results of Propensity-score Precision@k (PSP@k) are defer to Appendix~\ref{apx:psp},
which focus more on tail labels' performance.

\begin{table*}[h]
    \centering
    \caption{Data statistics. $N_{train}, N_{test}$ refer to the
      number of instances in the training and test sets, respectively.
      $L$: the number of labels. $\bar{L}$: the average number of positive labels per
      instance. $\bar{n}$: average number of instances per label. $d_{tfidf}$:
      the sparse feature dimension of $\Phi_{tfidf}(\cdot)$.
      These six publicly available benchmark datasets, including the sparse
      TF-IDF features are downloaded from \url{https://github.com/yourh/AttentionXML} which
      are the same as \attentionxml~\citep{you2019attentionxml}
      \xtransformer~\cite{chang2020xmctransformer} and
      \lightxml~\cite{jiang2021lightxml} for fair comparison.
    }
	\label{table:data}
    \resizebox{1.0\textwidth}{!}{
    \begin{tabular}{c|rrrrrr}
    	\toprule
		Dataset           & $N_{train}$     & $N_{test}$       & $L$       & $\bar{L}$ & $\bar{n}$ & $d_{tfidf}$     \\
    	\midrule
        \eurlex         &       15,449    &           3,865  &     3,956 &     5.30  &     20.79 &  186,104\\
        \wikis          &       14,146    &           6,616  &    30,938 &    18.64  &      8.52 &  101,938\\
        \amzcat         &    1,186,239    &         306,782  &    13,330 &     5.04  &    448.57 &  203,882\\
        \wikil          &    1,779,881    &         769,421  &   501,070 &     4.75  &     16.86 &2,381,304\\
        \amzsmall       &      490,449    &         153,025  &   670,091 &     5.45  &      3.99 &  135,909\\
        \amzlarge       &    1,717,899    &         742,507  & 2,812,281 &    36.04  &     22.02 &  337,067\\
    	\bottomrule
    \end{tabular}
    }
\end{table*}

\paragraph{Baseline Methods.}
We compare \xrtransformer with state-of-the-art (SOTA) \xmc methods:
\annexml~\citep{tagami2017annexml},
\discmec~\citep{babbar2017dismec}, \pfastrexml~\citep{jain2016extreme},
\parabel~\citep{prabhu2018parabel}, \xtext~\citep{wydmuch2018no},
\bonsai~\citep{khandagale2019bonsai}, \xmlcnn~\citep{liu2017deep},
\xrlinear~\cite{yu2020pecos}, \attentionxml~\cite{you2019attentionxml},
\xtransformer~\cite{chang2020xmctransformer} and
\lightxml~\cite{jiang2021lightxml}.
We obtain most baseline results
from~\cite[Table 3]{you2019attentionxml} and~\cite[Table 3]{yu2020pecos}
except for the latest deep learning based algorithms~\citep{you2019attentionxml,chang2020xmctransformer,jiang2021lightxml}.
To have fair comparison on training time,
we use the same hardware (i.e., AWS p3.16xlarge) and the same inputs
(i.e., raw text, vectorized features, data split)
to obtain the results of \attentionxml, \xtransformer and \lightxml.
The hyper-parameter of \xrtransformer and more empirical results are
included in Appendix~\ref{sec:hypterparameter}.

\begin{table*}[t]
	\centering
    \caption{Comparison of \xrtransformer with recent \xmc methods on six
    public datasets. Results with a trailing reference are taken
    from~\cite[Table 3]{you2019attentionxml} and~\cite[Table 3]{yu2020pecos}.
    We obtain the results of $\text{\attentionxml}^*$, $\text{\lightxml}^*$,
    $\text{\xtransformer}^*$ and $\text{\xrtransformer}^*$ on the
    same vectorized feature matrix provided in~\cite{you2019attentionxml}.
    Due to GPU memory constraint, \lightxml is not able to run on \amzlarge.
	The PSP@k results are available in Appendix~\ref{apx:psp}.
    }
    \resizebox{1.0\textwidth}{!}{
    \begin{tabular}{c|ccc|ccc|ccc}
        \toprule
		Methods & P@1 & P@3 & P@5 & P@1 & P@3 & P@5 & P@1 & P@3 & P@5\\
        \hline
        \hline
		 & \multicolumn{3}{c}{ \eurlex } & \multicolumn{3}{c}{ \wikis } & \multicolumn{3}{c}{ \amzcat } \\
        \midrule
		\annexml~\citep{tagami2017annexml}       & 79.66 & 64.94 & 53.52 & 86.46 & 74.28 & 64.20 & 93.54 & 78.36 & 63.30 \\
		\discmec~\citep{babbar2017dismec}        & 83.21 & 70.39 & 58.73 & 84.13 & 74.72 & 65.94 & 93.81 & 79.08 & 64.06 \\
		\pfastrexml~\citep{jain2016extreme}      & 73.14 & 60.16 & 50.54 & 83.57 & 68.61 & 59.10 & 91.75 & 77.97 & 63.68 \\
		\parabel~\citep{prabhu2018parabel}       & 82.12 & 68.91 & 57.89 & 84.19 & 72.46 & 63.37 & 93.02 & 79.14 & 64.51 \\
		\xtext~\citep{wydmuch2018no}             & 79.17 & 66.80 & 56.09 & 83.66 & 73.28 & 64.51 & 92.50 & 78.12 & 63.51 \\
		\bonsai~\citep{khandagale2019bonsai}     & 82.30 & 69.55 & 58.35 & 84.52 & 73.76 & 64.69 & 92.98 & 79.13 & 64.46 \\
		\xmlcnn~\citep{liu2017deep}              & 75.32 & 60.14 & 49.21 & 81.41 & 66.23 & 56.11 & 93.26 & 77.06 & 61.40 \\
		\xrlinear~\cite{yu2020pecos}             & 84.14 & 72.05 & 60.67 & 85.75 & 75.79 & 66.69 & 94.64 & 79.98 & 64.79 \\
        \midrule
		$\text{\attentionxml}^*$                 & 86.93 & 74.12 & 62.16 & 87.34 & 78.18 & 69.07 & 95.84 & 82.39 & 67.32 \\
		$\text{\xtransformer}^*$                 & 87.61 & 75.39 & 63.05 & 88.26 & 78.51 & 69.68 & 96.48 & 83.41 & 68.19 \\
      $\text{\lightxml}^*$                     & 87.15 & 75.95 & \best{63.45} & \best{89.67} & 79.06 & 69.87 & 96.77 & \best{83.98} & \best{68.63} \\
      $\text{\xrtransformer}^*$                & \best{88.41} & \best{75.97} & 63.18 & 88.69 & \best{80.17} & \best{70.91} & \best{96.79} & 83.66 & 68.04 \\
        \hline
        \hline
         & \multicolumn{3}{c}{ \wikil } & \multicolumn{3}{c}{ \amzsmall }     & \multicolumn{3}{c}{ \amzlarge } \\
        \midrule
		\annexml~\citep{tagami2017annexml}       & 64.22 & 43.15 & 32.79 & 42.09 & 36.61 & 32.75 & 49.30 & 45.55 & 43.11 \\
		\discmec~\citep{babbar2017dismec}        & 70.21 & 50.57 & 39.68 & 44.78 & 39.72 & 36.17 & 47.34 & 44.96 & 42.80 \\
		\pfastrexml~\citep{jain2016extreme}      & 56.25 & 37.32 & 28.16 & 36.84 & 34.23 & 32.09 & 43.83 & 41.81 & 40.09 \\
		\parabel~\citep{prabhu2018parabel}       & 68.70 & 49.57 & 38.64 & 44.91 & 39.77 & 35.98 & 47.42 & 44.66 & 42.55 \\
		\xtext~\citep{wydmuch2018no}             & 65.17 & 46.32 & 36.15 & 42.54 & 37.93 & 34.63 & 42.20 & 39.28 & 37.24 \\
		\bonsai~\citep{khandagale2019bonsai}     & 69.26 & 49.80 & 38.83 & 45.58 & 40.39 & 36.60 & 48.45 & 45.65 & 43.49 \\
		\xmlcnn~\citep{liu2017deep}              &   -   &   -   &   -   & 33.41 & 30.00 & 27.42 &   -   &   -   &   -   \\
		\xrlinear~\cite{yu2020pecos}             & 65.59 & 46.72 & 36.46 & 43.38 & 38.40 & 34.77 & 47.40 & 44.15 & 41.87 \\
        \midrule
        $\text{\attentionxml}^*$                 & 76.74 & 58.18 & 45.95 & 47.68 & 42.70 & 38.99 & 50.86 & 48.00 & 45.82 \\
        $\text{\xtransformer}^*$                 & 77.09 & 57.51 & 45.28 & 48.07 & 42.96 & 39.12 & 51.20 & 47.81 & 45.07 \\
        $\text{\lightxml}^*$                     & 77.89 & 58.98 & 45.71 & 49.32 & 44.17 & 40.25 &   -   &   -   &   -   \\
        $\text{\xrtransformer}^*$                & \best{79.40} & \best{59.02} & \best{46.25} & \best{50.11} & \best{44.56} & \best{40.64} & \best{54.20} & \best{50.81} & \best{48.26} \\
        \bottomrule
    \end{tabular}
    }
    \label{table:main-results}
\end{table*}

\begin{table*}[t]
    \centering
	\caption{Comparing training time (in hours) of DNN-based methods that produce
	the SOTA results in Table~\ref{table:main-results}.
    The number following the model indicates the number of ensemble models used.
    }
    \resizebox{1.0\textwidth}{!}{
    \begin{tabular}{crrrr}
        \toprule
        Dataset & \attentionxml-3 & \xtransformer-9 & \lightxml-3 & \xrtransformer-3\\
        \hline
        \eurlex & 0.9           & 7.5           & 16.9      & \best{0.8}\\
        \wikis  & \best{1.5}    & 14.1          & 26.9      & \best{1.5}\\
        \amzcat & 24.3          & 147.6         & 310.6     & \best{13.2}\\
        \wikil  & \best{37.6}   & 557.1         & 271.3     & 38.0\\
        \amzsmall & 24.2        & 514.8         & 159.0     & \best{10.5}\\
        \amzlarge & 54.8        & 542.0         & -         & \best{29.3}\\
        \bottomrule
    \end{tabular}
    }
    \label{table:time-results}
\end{table*}

\begin{table*}[h]
    \centering
    \caption{Single model comparison of DNN based \xmc models.
    Training time on p3.16xlarge with 8 Nvidia V100 GPUs $T^8_{train}$ are reported for
    \attentionxml, \xtransformer and \xrtransformer, whereas time on single
    Nvidia V100 GPU $T^1_{train}$ is reported for \lightxml and \xrtransformer.
    }
    \resizebox{0.75\textwidth}{!}{
    \begin{tabular}{cccrrrr}
        \toprule
        Dataset & Method & P@1 & P@3 & P@5 & $T^1_{train}$ & $T^8_{train}$\\
        \hline
        \multirow{4}{*}{\wikis}   & \attentionxml-1 & 87.1 & 77.8 & 68.8 & - & \best{0.5}\\
                                  & \xtransformer-1 & 87.5 & 77.2 & 67.1 & - & 3.5\\
                                  & \lightxml-1     & 87.8 & 77.3 & 68.0 & 6.7 & -\\
                                  & \xrtransformer-1& \best{88.0} & \best{78.7}
                                  & \best{69.1} & \best{1.3} & \best{0.5} \\
        \midrule
        \multirow{4}{*}{\wikil}& \attentionxml-1 & 75.1 & 56.5 & 44.4 & - & \best{12.5}\\
                                  & \xtransformer-1 & 44.8 & 40.1 & 34.6 & - & 56.0\\
                                  & \lightxml-1     & 76.3 & 57.3 & 44.2 & 89.6 & -\\
                                  & \xrtransformer-1& \best{78.1} & \best{57.6}
                                  & \best{45.0} & \best{29.2} & \best{12.5}\\
        \midrule
        \multirow{4}{*}{\amzsmall}& \attentionxml-1 & 45.7 & 40.7 & 36.9 & - & 8.1\\
                                  & \xtransformer-1 & 44.8 & 40.1 & 34.6 & - & 56.0\\
                                  & \lightxml-1     & 47.3 & 42.2 & 38.5 & 53.0 & -\\
                                  & \xrtransformer-1& \best{49.1} & \best{43.8}
                                  & \best{40.0} & \best{8.1} & \best{3.4}\\
        \bottomrule
    \end{tabular}
  }
  \label{table:single-results}
\end{table*}

\paragraph{Model Performance.}
The comparisons of Precision@k (P@k) and training time are shown in
Table~\ref{table:main-results} and Table~\ref{table:time-results},
respectively.
The proposed \xrtransformer follows \attentionxml and \lightxml to use \textit{an ensemble of 3} models,
while \xtransformer uses \textit{an ensemble of 9} models~\citep{chang2020xmctransformer}.
More details about the ensemble setting can be found in Appendix~\ref{sec:hypterparameter}.
The proposed \xrtransformer framework achieves new SOTA results in
\textbf{14 out of 18} evaluation columns (combination of datasets and P@k),
and outperforms competitive methods on the large datasets.
Next, we show the training time of \xrtransformer is
significantly less than other DNN-based models.

\paragraph{Training Cost.}
Table~\ref{table:time-results} shows the training time for these DNN-based models.
To have fair comparison, all the experiments are conducted with float32
precision on AWS p3.16xlarge instance with 8 Nvidia V100 GPUs except for
\lightxml, which was run on single V100 GPU since multi-GPU
training is not implemented.
\xrtransformer consumes significantly less training time compared
with other transformer based models and the shallow BiLSTM model \attentionxml.
On Amazon-3M, \xrtransformer has \textit{20x} speedup over
\xtransformer while achieving even better P@k.
Finally, in table~\ref{table:single-results}, we compare
\xrtransformer with \lightxml under the single model setup (no ensemble), where
\xrtransformer still consistently outperforms \lightxml in P@k and training time.

\begin{table*}[h]
	\centering
    \caption{Comparing \xrtransformer with \pretrained and \wordvec embeddings 
    concatenated with \tfidf features. 
    }
    \resizebox{0.9\textwidth}{!}{
    \begin{tabular}{c|ccc|ccc|ccc}
        \toprule
		Methods & P@1 & P@3 & P@5 & P@1 & P@3 & P@5 & P@1 & P@3 & P@5\\
        \hline
        \hline
		    & \multicolumn{3}{c}{ \eurlex } & \multicolumn{3}{c}{ \wikis } & \multicolumn{3}{c}{ \amzcat } \\
        \midrule
		  $\text{\tfidf}$                 & 84.14 & 72.05 & 60.97 & 85.75 & 75.79 & 66.69 & 94.64 & 79.98 & 64.79 \\
		  $\text{\wordvec+\tfidf}$        & 84.35 & 71.27 & 59.10 & 86.11 & 76.92 & 66.45 & 94.53 & 79.44 & 63.94 \\
      $\text{\pretrained+\tfidf}$     & 84.92 & 71.40 & 59.36 & 85.78 & 78.30 & 68.33 & 95.05 & 80.12 & 64.53 \\
      $\text{\xrtransformer}$         & \best{88.41} & \best{75.97} & \best{63.18} & \best{88.69} & \best{80.17} & \best{70.91} & \best{96.79} & \best{83.66} & \best{68.04} \\
        \hline
        \hline
         & \multicolumn{3}{c}{ \wikil } & \multicolumn{3}{c}{ \amzsmall }     & \multicolumn{3}{c}{ \amzlarge } \\
        \midrule
      $\text{\tfidf}$                 & 65.59 & 46.72 & 36.46 & 43.38 & 38.40 & 34.77 & 47.40 & 44.15 & 41.87 \\
      $\text{\wordvec+\tfidf}$        & 68.21 & 48.16 & 37.54 & 44.04 & 39.07 & 35.35 & 47.51 & 44.49 & 42.19 \\
      $\text{\pretrained+\tfidf}$     & 70.18 & 49.82 & 38.75 & 44.55 & 38.91 & 34.77 & 49.66 & 46.41 & 43.96 \\
      $\text{\xrtransformer}$         & \best{79.40} & \best{59.02} & \best{46.25} & \best{50.11} & \best{44.56} & \best{40.64} & \best{54.20} & \best{50.81} & \best{48.26} \\
        \bottomrule
    \end{tabular}
    }
    \label{table:embedding-compare}
\end{table*}

\paragraph{Comparison of Different Semantic Embeddings.}
To provide more empirical justifications, that the improvement in performance 
comes from better semantic embedding rather than the introducing of \tfidf 
features, we further tested models using 
\pretrained Transformer and \wordvec embeddings concatenated with the same 
\tfidf features. 

Table~\ref{table:embedding-compare} summarizes the performance of these models 
on all 6 datasets. In particular, \wordvec is using token embedding from 
\textit{word2vec-google-news-300} and for \pretrained we use the same setting as 
\xrtransformer (3-model ensemble). On large datasets 
such as \wikil/\amzsmall/\amzlarge, \pretrained+\tfidf has marginal 
improvement compared to the baseline \tfidf features. Nevertheless, our 
proposed \xrtransformer still enjoy significant gain compared to 
\pretrained+\tfidf. This suggests the major improvement is from learning more 
powerful neural semantic embeddings, rather than the use of \tfidf. 

\paragraph{Effect of Cost-Sensitive Learning.}
In Table~\ref{table:cost-sensitive-ablation},
we analyze the effect of cost sensitive learning on four \xmc datasets with 
the largest output spaces.
On most datasets, cost sensitive learning via aggregated labels yields better performance than
those without. We also show that cost-sensitive learning is not only beneficial to \xrtransformer,
but also useful to its linear counterpart \xrlinear~\citep{yu2020pecos}.
See Appendix~\ref{apx:cost-sensitive} for more results.
\begin{table*}[h]
	\centering
	\caption{Ablation of cost-sensitive learning on the single
		\xrtransformer model with or without Cost Sensitive~(CS).
		Precision@1,3,5 P(@k) and Recall@1,3,5 (R@k) are reported.
	}
    \resizebox{0.9\textwidth}{!}{
    \begin{tabular}{ll|rrr|rrr}
        \toprule
		Dataset 					& Method 		& P@1  & P@3  & P@5  & R@1  & R@3  & R@5  \\
        \hline
		\multirow{2}{*}{\wikis}		
                  & \xrtransformer-1(w/o CS) 	& 86.8 & 77.6 & 68.8 & 5.2 & 13.6 & 19.8 \\
                  & \xrtransformer-1 	        & 88.0 & 78.7 & 69.1 & 5.3 & 13.8 & 19.9 \\
    	\midrule
		\multirow{2}{*}{\wikil}		
                  & \xrtransformer-1(w/o CS) 	& 77.6 & 57.4 & 44.9 & 25.8 & 48.1 & 57.8 \\
                  & \xrtransformer-1 	        & 78.1 & 57.6 & 45.0 & 26.1 & 48.5 & 58.1 \\
	    \midrule
		\multirow{2}{*}{\amzsmall}	
                  & \xrtransformer-1(w/o CS) 	& 49.1 & 43.8 & 40.0 & 10.3 & 25.6 & 37.7 \\
                  & \xrtransformer-1 	        & 49.0 & 43.7 & 39.9 & 10.4 & 25.7 & 37.7 \\
	    \midrule
		\multirow{2}{*}{\amzlarge}	
                  & \xrtransformer-1(w/o CS) 	& 50.2 & 47.6 & 45.4 &  3.4 &  8.4 & 12.4 \\
                  & \xrtransformer-1 	        & 52.6 & 49.4 & 46.9 &  3.8 &  9.3 & 13.6 \\
		\bottomrule
    \end{tabular}
  	}
    \label{table:cost-sensitive-ablation}
\end{table*}

\begin{wrapfigure}{r}{0.55\textwidth}
\centering
\includegraphics[width=0.55\textwidth]{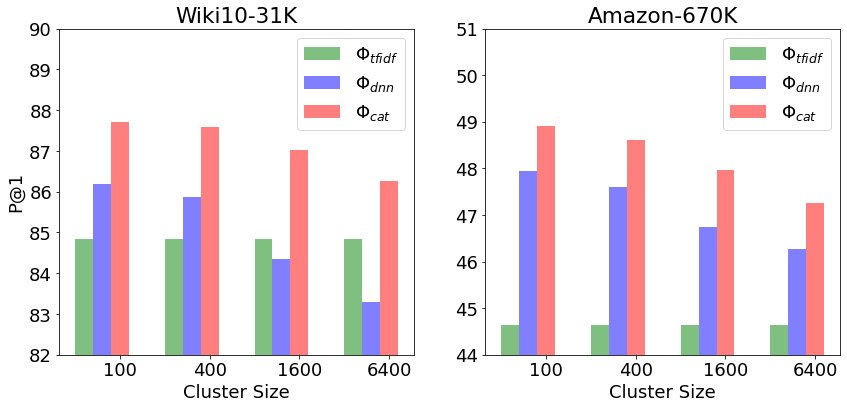}
  \caption{Comparison of BERT model fine-tuned with different label
  resolution. Larger cluster size means lower label resolution. Note that 
  $\Phi_{cat}$ is the normalized concatenation of $\Phi_{tfidf}$ and 
  $\Phi_{dnn}$.}
  \label{fig:cluster_size}
\end{wrapfigure}

\paragraph{Effect of Label Resolution and Text Representation.}
Next, we compare the effect of label resolution on the quality of the
fine-tuned transformer embeddings. We
fine-tune transformer models in a non-recursive manner on a two layer HLT with
different leaf cluster size. Then the fine-tuned transformer embeddings are
used along or in combination with TF-IDF features to produce the predictions
with refined HLT. From Figure~\ref{fig:cluster_size} we can observe that a
larger cluster size will result in worse semantic features.
Figure~\ref{fig:cluster_size} also shows that combining semantic features
$\Phi_{dnn}$ with statistical features $\Phi_{tfidf}$ could in general improve
the model performance.

\section{Conclusion and Future work}
\label{sec:conclusion}
In this paper, we have presented \xrtransformer approach, which is an XMC architecture
that leverages multi-resolution objectives and cost sensitive learning to 
accelerate the fine-tuning of pre-trained transformer models.
Experiments show that the proposed method 
establishes new state-of-the-art results on public XMC datasets while taking 
significantly less training time compared with earlier transformer based 
methods. Although the proposed architecture is designed for XMC, the ideas
can be applied to other areas such as information retrieval or other DNN 
models such as CNNs/ResNets. 
Also, more extensive study is required to understand why the coarse-to-fine scheme 
would lead to not only faster training but better overall quality.
A hypothesis is that the problem is being solved at multiple scales
hence leading to more robust learning of deep transformer models.

\newpage
\bibliographystyle{unsrt}
\bibliography{midas}

\newpage
\appendix
\section{Appendix}
\subsection{Method Overview}
\begin{figure}[h]
\centering
\includegraphics[width=1.0\textwidth]{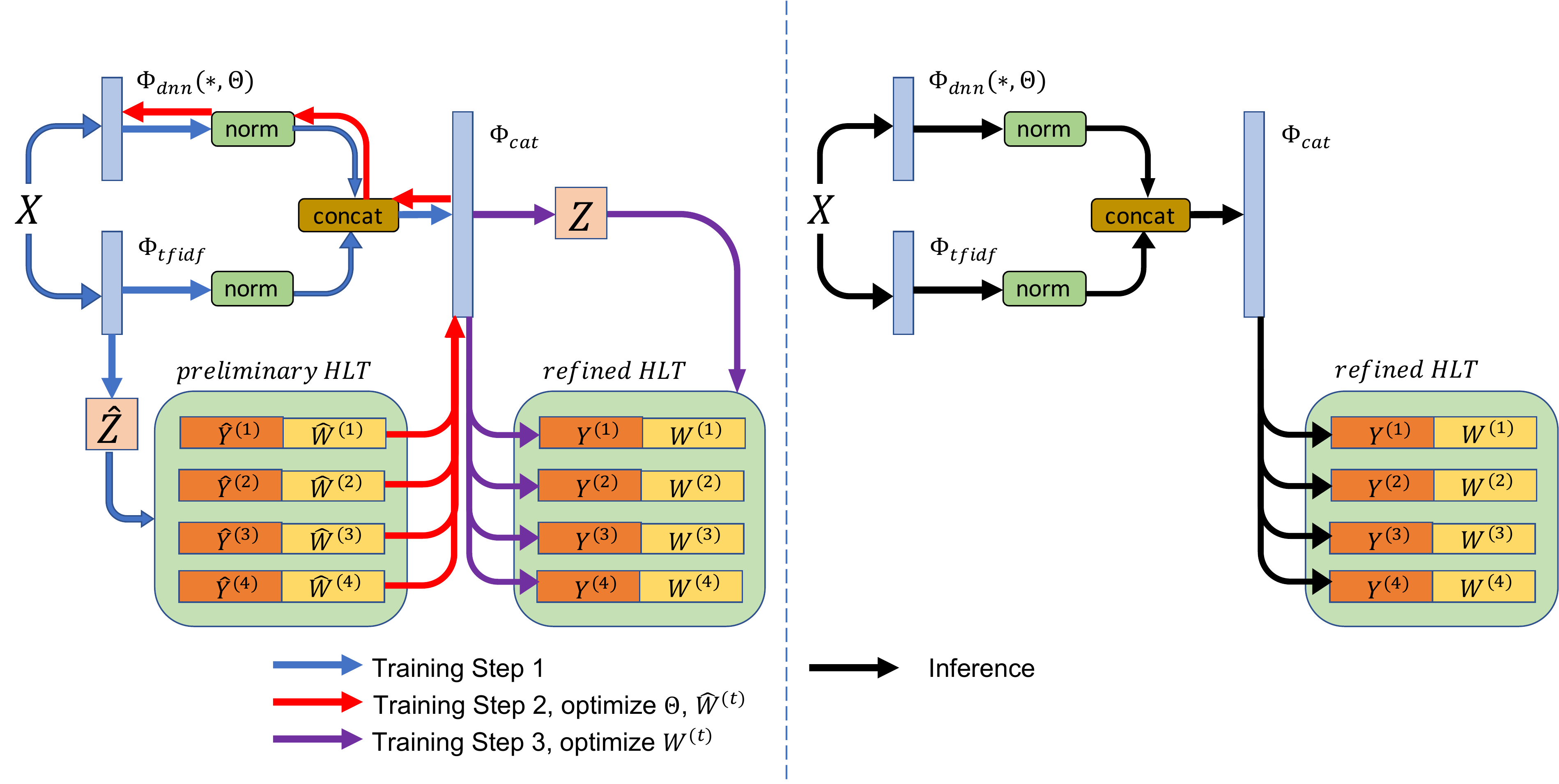}
  \caption{\xrtransformer training~(left) and inference~(right) architecture.
  \xrtransformer is trained with three steps.  First, label
  features $\hat{\Zb}$ are computed and is used to build preliminary
  hierarchical label tree~(HLT) via hierarchical k-means.
  Then the transformer vectorizer $\Phi_{dnn}(\cdot,\Theta)$ is recursively
  fine-tuned on multi-resolution labels $\{\hat{\Yb}^{(t)}\}_{t=1}^D$.
  Finally, a refined HLT is generated with $\Phi_{cat}$ and
  the linear ranking models $\{\Wb^{(t)}\}_{t=1}^D$ are learned with refined
  multi-resolution labels $\{\Yb^{(t)}\}_{t=1}^D$. Only the transformer
  vectorizer $\Phi_{dnn}$, refined HLT and $\{\Wb^{(t)}\}_{t=1}^D$ are needed
  during inference.}
  \label{fig:xrtransformer_arch}
\end{figure}

At a high level sketch,
the \xrtransformer model consists of three components. First, text vectorizers
consists of both learnable transformer based semantic vectorizer
and deterministic statistical vectorizer.
Second, a recursive fine-tuning curriculum with multi-resolution output
signals defined by a preliminary HLT and finally, multi-layer ranking models that gives
final prediction together with a refined HLT. 
Figure~\ref{fig:xrtransformer_arch} gives an illustration on the training and 
inference pipeline of \xrtransformer. 

\subsection{Evaluation Metrics}
In this section, we define the evaluation metrics used in this paper.
The most widely used evaluation metric for \xmc is the
precision at k~(Prec@k) and recall at k~(Recall@k), which are defined as:
\begin{align}
  \text{Prec@k}&=\frac{1}{k}\sum_{l=1}^k y_{rank(l)}\\
  \text{Recall@k}&=\frac{1}{\sum_{i=1}^L y_i}\sum_{l=1}^k y_{rank(l)}
\end{align}
where $y\in\{0,1\}^L$ is the ground truth label and $rank(l)$ is the index of the $l$-th highest predicted label.

For performance comparison on tail labels, we also report propensity scored
precision~(PSP@k), which is defined as:
\begin{align}
  \text{PSP@k}= \frac{1}{k}\sum_{l=1}^k \frac{y_{rank(l)}}{p_{rank(l)}}
\end{align}
where $p_{rank(l)}$ is the propensity score at position $rank(l)$~\cite{jain2016extreme}.
The metric involves application specific parameters $A$ and $B$. For
consistency, we use the same setting as
\attentionxml~\cite{you2019attentionxml} for all datasets.

\subsection{Model Hyperparameters}
\label{sec:hypterparameter}
For each sub-XMC problem, \xrtransformer fine-tunes the transformer encoder
with Adam~\cite{kingma2014adam} with linear learning rate schedule and a batch
size of 32 per GPU~(256 total). The
dropout probability is set to be 0.1 for all models and
bootstrapping~\eqref{eq:obj_bootstrapping} is used for all fine-tuning
problems except for the root layer. The detailed hyperparameters are listed in Table~\ref{table:hyperparameters}.

\subsubsection{Model Ensemble Setup}
For \attentionxml, \lightxml and \xtransformer, we use the same ensemble setting provided in the paper. In
particular, \attentionxml uses ensemble of 3 models, \lightxml uses
ensemble of 3 transformer encoders and \xtransformer
uses 9 model ensemble with BERT, RoBerta, XLNet large models with three
difference indexers. For our method, we follow the setting of \lightxml and
use the ensemble of BERT, RoBerta and XLNet for \eurlex, \wikis and \amzcat, 
and ensemble of three BERT model for \wikil, \amzsmall and \amzlarge.

\begin{table*}[h]
    \centering
    \caption{Hyperparameters for \xrtransformer. $HLT_{prelim}$ and
    $HLT_{refine}$ define the structures of the preliminary and refined
    hierarchical label trees. $lr_{max}$ is the maximum learning rate used in
    fine-tuning. $n_{step}$ is the total number of optimization steps across
    the multi-resolution fine-tuning. $N_x$ is the number of input tokens after
    text truncation. $\alpha$ is the hyper-parameter for cost sensitive
    learning. $\lambda$ is the weight for the regularization term.}
    \label{table:hyperparameters}
	\resizebox{1.0\textwidth}{!}{
    \begin{tabular}{c|rrrrrrrr}
    \toprule
      Dataset     & $HLT_{prelim}$  & $HLT_{refine}$   & $lr_{max}$ & $n_{step}$& $N_x$ & $\alpha$ & $\lambda$\\
    \midrule
        \eurlex   &     16-256-3956 &   4-32-256-3956  &  $5\times 10^{-5}$ &  2400 & 128  & 1.0 & 0.5\\
        \wikis    &  128-2048-30938 & 8-128-2048-30938 &  $1\times 10^{-4}$ &  4000 & 256  & 0.25 & 0.25\\
        \amzcat   &  128-1024-13330 &   128-1024-13330 &  $1\times 10^{-4}$ & 45000 & 256  & 2.0 & 2.0\\
        \wikil    &64-512-4096-32768&8-256-8192-501070 &  $1\times 10^{-4}$ & 60000 & 128  & 0.25 & 0.25\\
        \amzsmall &   128-2048-32768&8-256-8192-670091 &  $1\times 10^{-4}$ & 20000 & 128  & -- & 1.0\\
        \amzlarge &   128-2048-32768&8-256-8192-2812281&  $1\times 10^{-4}$ & 30000 & 128  & 0.125 & 0.125\\
    \bottomrule
    \end{tabular}
    }
\end{table*}

\subsection{More Empirical Results}

\begin{table*}[h]
    \centering
	\caption{Comparing training time (in hours) of DNN-based methods that produce
	the SOTA results in Table~\ref{table:main-results}.
    The number following the model indicates the number of ensemble models used.
    }
    \resizebox{1.0\textwidth}{!}{
    \begin{tabular}{crrrr}
        \toprule
        Dataset & \attentionxml-3 & \xtransformer-9 & \lightxml-3 & \xrtransformer-3\\
        \hline
        \eurlex & 0.9           & 7.5           & 16.9      & \best{0.8}\\
        \wikis  & \best{1.5}    & 14.1          & 26.9      & \best{1.5}\\
        \amzcat & 24.3          & 147.6         & 310.6     & \best{13.2}\\
        \wikil  & \best{37.6}   & 557.1         & 271.3     & 38.0\\
        \amzsmall & 24.2        & 514.8         & 159.0     & \best{10.5}\\
        \amzlarge & 54.8        & 542.0         & -         & \best{29.3}\\
        \bottomrule
    \end{tabular}
    }
    \label{table:time-results}
\end{table*}

\subsubsection{Cost-sensitive Learning}
\label{apx:cost-sensitive}
In table~\ref{table:cost-sensitive-all}, we show that
the cost-sensitive learning via recursive label aggregation
is beneficial to both the linear \xrlinear and the Transformer-based \xrtransformer model.
On most datasets, cost sensitive learning yields better performance.

\begin{table*}[h]
    \centering
	\caption{Ablation of cost-sensitive learning on the single
		\xrlinear and \xrtransformer model with or without Cost Sensitive~(CS).
		Precision@1,3,5 (P@k) and Recall@1,3,5 (R@k) are reported.
	}
    \resizebox{1.0\textwidth}{!}{
    \begin{tabular}{ll|rrr|rrr}
        \toprule
		Dataset 					& Method 		& P@1  & P@3  & P@5  & R@1  & R@3  & R@5  \\
        \hline
		\multirow{4}{*}{\wikis}
				  & \xrlinear-1(w/o CS) 	    & 84.5 & 73.1 & 64.1 & 5.0 & 12.8 & 18.4 \\
				  & \xrlinear-1  	        	& 85.7 & 75.0 & 65.2 & 5.1 & 13.1 & 18.7 \\
                  & \xrtransformer-1(w/o CS) 	& 86.8 & 77.6 & 68.8 & 5.2 & 13.6 & 19.8 \\
                  & \xrtransformer-1 	        & 88.0 & 78.7 & 69.1 & 5.3 & 13.8 & 19.9 \\
    	\midrule
		\multirow{4}{*}{\wikil}
				  & \xrlinear-1(w/o CS)     	& 66.7 & 47.8 & 37.3 & 21.7 & 39.5 & 47.6 \\
				  & \xrlinear-1     	        & 67.7 & 48.4 & 37.6 & 22.2 & 40.4 & 48.6 \\
                  & \xrtransformer-1(w/o CS) 	& 77.6 & 57.4 & 44.9 & 25.8 & 48.1 & 57.8 \\
                  & \xrtransformer-1 	        & 78.1 & 57.6 & 45.0 & 26.1 & 48.5 & 58.1 \\
	    \midrule
		\multirow{4}{*}{\amzsmall}
				  & \xrlinear-1(w/o CS)		    & 44.0 & 39.2 & 35.4 &  9.2 & 22.7 & 33.2 \\
				  & \xrlinear-1	        		& 44.4 & 39.3 & 35.6 &  9.4 & 23.0 & 33.6 \\
                  & \xrtransformer-1(w/o CS) 	& 49.1 & 43.8 & 40.0 & 10.3 & 25.6 & 37.7 \\
                  & \xrtransformer-1 	        & 49.0 & 43.7 & 39.9 & 10.4 & 25.7 & 37.7 \\
	    \midrule
		\multirow{4}{*}{\amzlarge}
				  & \xrlinear-1(w/o CS)			& 46.8 & 44.0 & 41.9 &  2.9 &  7.3 & 10.7 \\
				  & \xrlinear-1		 	        & 50.1 & 46.6 & 44.0 &  3.4 &  8.4 & 12.2 \\
                  & \xrtransformer-1(w/o CS) 	& 50.2 & 47.6 & 45.4 &  3.4 &  8.4 & 12.4 \\
                  & \xrtransformer-1 	        & 52.6 & 49.4 & 46.9 &  3.8 &  9.3 & 13.6 \\
		\bottomrule
    \end{tabular}
  	}
    \label{table:cost-sensitive-all}
\end{table*}

\subsubsection{Inference Speed}
\label{apx:inference}
We report the inference time for \attentionxml, \xtransformer, \lightxml and 
\xrtransformer on \xmc datasets in table~\ref{table:inference-time}.
The inference results (millisecond per sample) are evaluated on single GPU and 
single CPU for 
most comparing models except for \attentionxml, which is evaluated with 
multi-GPUs on \wikil, \amzsmall and \amzlarge. \attentionxml requires 
model-parallelism on those largest datasets otherwise it may be out-of-memory 
on a single-GPU setup. 

\begin{table*}[h]
    \centering
    \caption{Comparison of \xrtransformer with recent \xmc methods on
    public datasets w.r.t. inference time. Times are recorded with single 
    Nvidia V100 GPU and batch size of 1 except for 
    the numbers with superscript $^*$, where model parallel was used and 
    inference was done with 8 GPUs.    
    The unit is milliseconds per sample.  
    }
    \resizebox{1.0\textwidth}{!}{
    \begin{tabular}{crrrr}
        \toprule
        Dataset & \attentionxml-1 & \xtransformer-1 & \lightxml-1 & \xrtransformer-1\\
        \hline
        \eurlex & 12.7          & 48.2          & 24.7      & 22.3\\
        \wikis  & 20.0          & 48.1          & 27.1      & 39.1\\
        \amzcat & 14.4          & 47.6          & 24.1      & 26.1\\
        \wikil  & 80.1$^*$      & 48.1          & 27.3      & 33.9\\
        \amzsmall & 76.0$^*$    & 48.0          & 23.3      & 30.9\\
        \amzlarge & 130.5$^*$   & 50.2          & -         & 35.2\\
        \bottomrule
    \end{tabular}
    }
    \label{table:inference-time}
\end{table*}

\subsubsection{Propensity-score Precision}
\label{apx:psp}
We report the propensity scored precision~(PSP@k) metric on large scale \xmc
datasets for \pfastrexml, \parabel, \attentionxml and \xrtransformer in
Table~\ref{table:psp-results}. For
consistency, we use the same setting as
\attentionxml~\cite{you2019attentionxml} for all datasets.

\begin{table*}[h]
    \centering
    \caption{Comparison of \xrtransformer with recent \xmc methods on 3 large
    public datasets w.r.t. PSP@k~(propensity scored precision at $k=1,3,5$).
    Results with a trailing reference are taken
    from~\cite[Table 6]{you2019attentionxml}.
    We obtain the results of $\text{\attentionxml}^*$
    and $\text{\xrtransformer}$ on the
    same vectorized feature matrix provided in~\cite{you2019attentionxml} and
    same hyper-parameter for propensity score calculation.
    }
    \resizebox{1.0\textwidth}{!}{
    \begin{tabular}{cccccccccc}
        \toprule
        & \multicolumn{3}{c}{ \wikil } & \multicolumn{3}{c}{ \amzsmall }     & \multicolumn{3}{c}{ \amzlarge } \\
        \hline
        \hline
        Methods & PSP@1 & PSP@3 & PSP@5 & PSP@1 & PSP@3 & PSP@5 & PSP@1 & PSP@3 & PSP@5\\
        \midrule
        \pfastrexml~\citep{jain2016extreme}      & 32.02 & 29.75 & 30.19 & 20.30 & 30.80 & 32.43 & \best{21.38} & 23.22 & 24.52 \\
        \parabel~\citep{prabhu2018parabel}       & 26.88 & 31.96 & 35.26 & 26.36 & 29.95 & 33.17 & 12.80 & 15.50 & 17.55 \\
        \midrule
        $\text{\attentionxml}^*$                 & 30.69 & 38.92 & 44.00 & 30.25 & 33.88 & 37.18 & 15.42 & 18.32 & 20.48 \\
        $\text{\xrtransformer}^*$                & \best{35.76} & \best{42.22} & \best{46.36} & \best{36.16} & \best{38.39} & \best{40.99} & 20.52 & \best{23.64} & \best{25.79} \\
        \bottomrule
    \end{tabular}
    }
	\label{table:psp-results}
\end{table*}

\subsection{Potential Negative Societal Impacts}
\label{sec:societal_impact}
This paper focus on the acceleration of the training algorithms on 
\xmc. To the best of our knowledge, our work poses no negative societal 
impacts.  

\end{document}